\documentclass[letterpaper, 10 pt, conference]{ieeeconf}  

\IEEEoverridecommandlockouts                              

\usepackage[space]{cite}

\usepackage{graphics} 
\usepackage{graphicx}
\usepackage{color}
\usepackage[percent]{overpic}
\usepackage{algorithmicx}
\usepackage{algpseudocode}
\usepackage{algorithm}

\usepackage{units}

\newcommand{\rom}[1]{\uppercase\expandafter{\romannumeral #1\relax}}

\algrenewcommand\textproc{\textbf}
\usepackage{booktabs}
\algtext*{EndFor}
\algtext*{Procedure}
\algtext*{EndProcedure}
\algtext*{EndIf}
\algtext*{EndWhile}
\usepackage{amsmath} 
\usepackage{amssymb}  

\usepackage[export]{adjustbox}
\usepackage{amsthm}
\theoremstyle{plain}
\newtheorem{theo}{Theorem}
\theoremstyle{plain}
\theoremstyle{plain}
\theoremstyle{plain}
\theoremstyle{plain}

\theoremstyle{definition}
\newtheorem{defn}{Definition}
\theoremstyle{definition}

\theoremstyle{definition}
\theoremstyle{definition}

\theoremstyle{remark}
\newtheorem{rema}[theo]{Remark}
\theoremstyle{remark}

\usepackage{lipsum}

\newcommand{\myx}{\mbox{{\fontfamily{ppl}\selectfont x}}}

\title{\LARGE \bf
Robust, Compliant Assembly via Optimal Belief Space Planning
}

\author{Florian Wirnshofer$^{1}$, Philipp S. Schmitt$^{1}$, Wendelin Feiten$^{1}$, Georg v. Wichert$^{1}$ and Wolfram Burgard$^{2}$
\thanks{$^{1}$Siemens Corporate Technology, Otto-Hahn-Ring 6, 81739 Munich,
	Germany}%
\thanks{$^{2}$Department of Computer Science, University of Freiburg, 79110
	Freiburg, Germany}%
}

\begin{document}

\maketitle
\thispagestyle{empty}
\pagestyle{empty}

\begin{abstract}

In automated manufacturing, robots must reliably assemble parts of various geometries and low tolerances. Ideally, they plan the required motions autonomously.
This poses a substantial challenge due to high-dimensional state spaces and non-linear contact-dynamics.
Furthermore, object poses and model parameters, such as friction, are not exactly known and a source of uncertainty.
The method proposed in this paper models the task of parts assembly as a belief space planning problem over an underlying impedance-controlled, compliant system.
To solve this planning problem we introduce an asymptotically optimal belief space planner by extending an optimal, randomized, kinodynamic motion planner to non-deterministic domains.
Under an expansiveness assumption we establish probabilistic completeness and asymptotic optimality.
We validate our approach in thorough, simulated and real-world experiments of multiple assembly tasks.
The experiments demonstrate our planner's ability to reliably assemble objects, solely based on CAD models as input.
\end{abstract}

\section{INTRODUCTION}
Industrial domains such as manufacturing or logistics require the automation of various processes that involve the joining of objects with robotic manipulators.
Examples include packaging or assembly. Fig.~\ref{fig:real_robot} shows two exemplary applications.
The state-of-the-art method of automating these processes is to rigidly and precisely fix objects in a structured environment and to hard-code all motions of the manipulator.
Often problem specific hardware such as grippers, fixtures and guiding components are designed for the task at hand. 

This approach is impractical for domains that demand short product life cycles and mass customization. 
The assembly task may even only occur once. 
The required degree of flexibility certainly cannot be achieved when relying on problem specific fixtures. We need an automated method that despite the increase in process uncertainties computes reliable and robust robot assembly motions.

A straightforward approach is to plan the geometric motions of the object that is to be assembled and then execute these motions on a compliant robot.
This simple approach has several drawbacks.
The dynamic execution of geometric paths, the uncertainty of the process and the non-linear contact-dynamics are not independent.
In order to compute highly robust and optimal motions, these aspects of the assembly problem must be considered simultaneously.
Practical, autonomous solutions for the assembly task must therefore address the following challenges:

\begin{figure}
	\centering
	\includegraphics[width=0.95\linewidth]{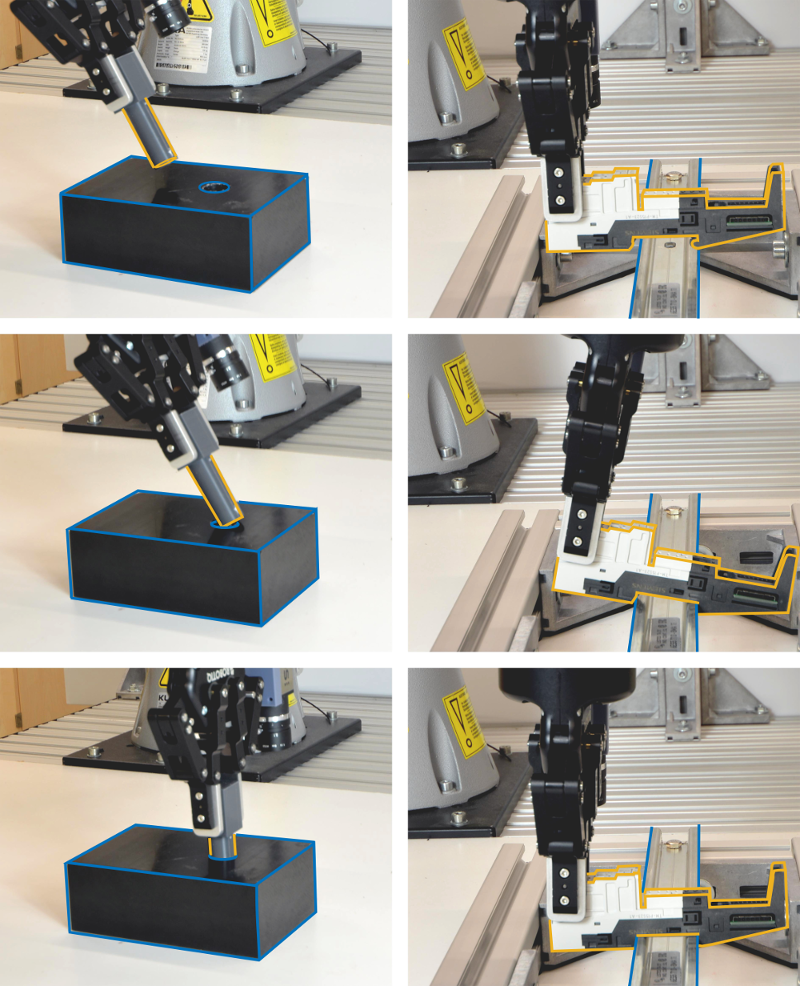}
	\caption{Two assembly tasks - An object (marked yellow) is attached to the robot and must be joined to a second object (marked blue) that is fixed to the environment.}
	\label{fig:real_robot}
\end{figure}

\subsubsection{High-dimensional and hybrid, non-linear dynamics}
Assembly requires the relative motion of two parts in up to six Cartesian degrees of freedom (DoF).
The process is dynamic which requires the inclusion of at least velocities, possibly further derivatives, into the configuration space.
As contacts play an essential role in assembly the dynamics exhibit abrupt non-linearities or hybrid dynamics.

\subsubsection{Uncertainties in model and process} 
Joining parts is subject to various sources of uncertainty.
The physical properties of the process at hand are difficult to obtain.
Parameters such as friction or elasticities may not be available at all.
Furthermore, poses of objects are only known with limited precision.

\subsubsection{Optimality}
To achieve or exceed human-level efficiency, the robot motions should be optimal with respect to the performance requirements of the task.
Key performance indicators for industrial assembly would be cycle times, reliability and low part breakage.
These performance indicators require the consideration of the full dynamics and uncertainty simultaneously.

The contribution of this paper is a planning method that addresses all of the above challenges.
We model the planning problem as a belief space planning problem over an underlying impedance-controlled robot.
We propose an asymptotically optimal belief space planner to solve this problem class.
This planner is implemented using a particle-based representation of uncertainty and a physics simulator.
Extensive experiments in simulation and on a real robot show that our method reliably joins parts with high-quality trajectories.

\section{RELATED WORK} \label{sec:related_work}

The work presented in this paper integrates and extends three strands of the planning literature:
compliant motions for parts assembly, optimal, kinodynamic motion planning and belief space planning.

\subsection{Compliant Motions for Assembly}

Compliance is a key requirement for successful parts assembly in realistic, uncertain environments. Fine motion planning \cite{lozano1984automatic} derives compliant motions by a backward chaining of so-called preimages. However, this approach does not scale to scenarios of practically relevant dimensionality \cite{canny1989computability}. Notable extensions seek to overcome the scalability issue by hierarchization over contact states \cite{dakin1992simplified} or by the application of sampling-based planners \cite{ji2001planning}. Yet, these techniques require a manual specification of contact states and are non-optimal. Model-based approaches towards parts assembly show convincing results in practice, yet require manual, problem specific modeling of motion constraints \cite{bruyninckx1995peg,DeSchutter-ijrr2007}. Methods from reinforcement learning \cite{gullapalli1994acquiring, levine2015learning, levine2016end} bypass the need of precise constraint modeling yet require the cumbersome design of cost functions.
%
%
%
%
Additionally the proposed methods require elaborate real-world experiments, in the case of \cite{gullapalli1994acquiring} more than one hundred iterations. 

\subsection{Optimal, Kinodynamic Motion Planning}

Planning for robotic manipulators occurs in high-dimensional configuration spaces in which sampling-based planners are proven to be highly efficient.
These include the probabilistic roadmap (PRM)~\cite{prm}, rapidly exploring random tree (RRT)~\cite{rrt} and the expansive space tree (EST)~\cite{hsu1997path}.
These planners require the availability of a local planner or steering function, which is trivial to obtain for geometric planning but potentially unavailable for domains with differential constraints.

To allow a more general formulation of the planning domains using a control-based model, kinodynamic extensions to the RRT have been proposed in~\cite{kinodynamic_rrt}.
In~\cite{est_moving_obstacles_icra} an extension to the EST planner is proposed that can handle both kinodynamic and time-varying domains.
The planner proposed in this paper is direct extension of this variant of the EST to belief space planning problems.

Sampling-based planners typically return jerky and far from optimal solutions.
In~\cite{karaman_main} a proof for the sub-optimality of PRM and RRT is provided along with asymptotically optimal counterparts PRM* and RRT*. These planners have been extended to kinodynamic planning~\cite{karaman_kinodynamic, karaman2013sampling, Li2016sst}.

In~\cite{hauser_aox} the AO-$x$ meta-planner is introduced.
This algorithm uses a probabilistically complete, kinodynamic motion planner repeatedly for optimal results in the limit.
We make use of the AO-$x$ algorithm to plan optimally in the limit in a belief space instead of a configuration space.

\subsection{Belief Space Planning for Assembly}
Planning under motion and sensing uncertainties is generally referred to as belief space planning. Exact solutions to the corresponding Partially-Observable Markov Decision Process (POMDP) only exist for a very limited problem class. The assumption of Gaussian process- and measurement noise \cite{platt2010belief, bry2011rapidly, van2011lqg}  can provide a remedy regarding computational complexity. However, these assumptions tend to be inadequate in the face of non-linear, hybrid dynamics as encountered in parts assembly. Further works \cite{melchior2007particle, hauser2010randomized} apply RRTs to belief space planning problems. RRTs are suboptimal and require a steering function, which for our application is difficult to obtain.

Recent works focusing on parts assembly \cite{kim2017, phillipsplanning, sieverling2017interleaving}, employ sampling-based belief approximations. Robot and environment uncertainty are represented using particles. As is the case for our work, a physics simulator is used in order to integrate each particle's system dynamics. In contrast to our work, Contact-Exploiting RRT \cite{sieverling2017interleaving} requires the distinction between free-space and contact motions. The work of \cite{kim2017} relies on a discretization of the action space as a fixed set of motion primitives. Neither of these methods achieve optimal solutions.

\section{Problem Statement and Notation} \label{sec:problem_statement}

We consider the problem of joining two parts modeled as rigid bodies.
One of the parts is rigidly attached to a robotic manipulator.
A second part is rigidly attached to the robot's environment. Object poses relative to the manipulator or environment are not exactly known.
The goal of the robot is to bring both parts into a desired relative pose.

In Section~\ref{subsec:compliance} we introduce a corresponding belief space formulation. We state our notation for a generalized kinodynamic, belief space planning problem in Section~\mbox{\ref{subsec:belief}}.

\subsection{Compliant Environment Interaction under Uncertainty}\label{subsec:compliance}

\begin{figure*}[tb]
\vspace{0.1cm}
\centering
\begin{overpic}[width=0.93\textwidth]{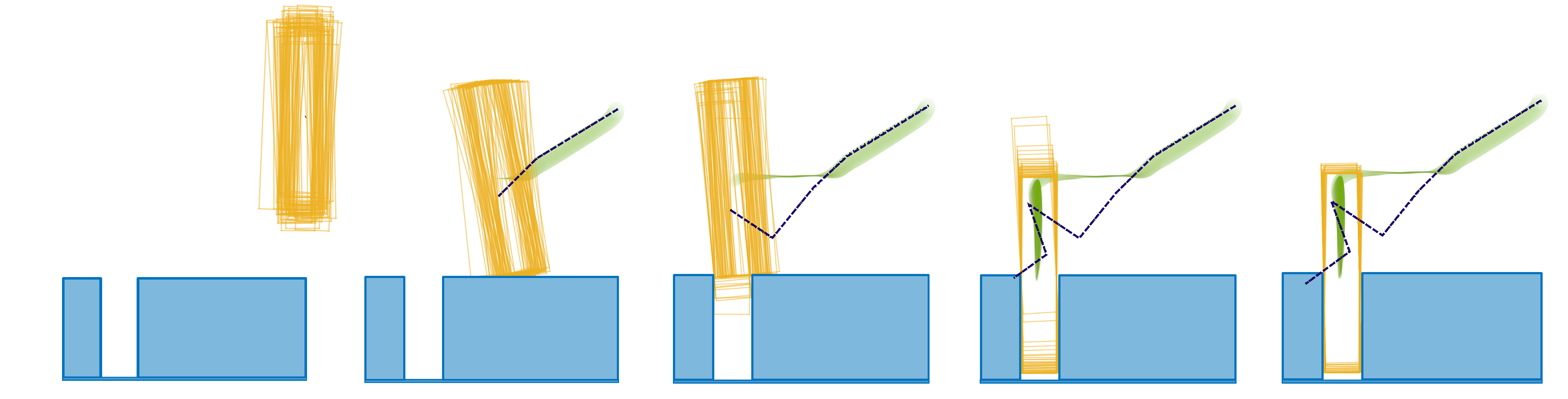}
	\put(12,-1.00){\rom{1}}
	\put(31,-1.00){\rom{2}}
	\put(50,-1.00){\rom{3}}
	\put(70,-1.00){\rom{4}}
	\put(90,-1.00){\rom{5}}
	\put(90,-1.00){\rom{5}}
	\put(13,21.85){\vector(2,-1){3}}
	\put(39,21.85){\vector(-1,-2){2.5}}
	\put(39,24){\scriptsize{set-point}}
	\put(38.6,22.4){\scriptsize{trajectory}}
	\put(8,24){\scriptsize{object pose}}
	\put(7.8,22.4){\scriptsize{distribution}}
\end{overpic}
\caption{Generating a robust set-point trajectory (dark line) for the peg-in-hole insertion task. 
The yellow boxes indicate a particle representation of the distribution of peg poses.
Each peg follows the set point via the spring damper dynamics of a Cartesian impedance controller.
The green area visualizes the translational covariance of peg poses.}
\label{fig:figure_trajectory}
\vspace{-0.4cm}
\end{figure*}
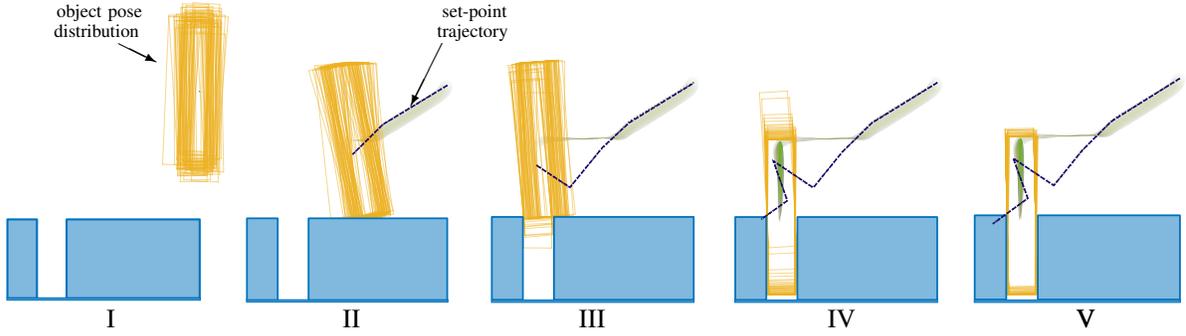

During the assembly task contact situations may occur in which the environment geometry sets constraints on the paths the robot can follow. When using stiff, position-controlled robots, even minor modeling inaccuracies or process uncertainties can build up contact forces that are far beyond the physical capabilities of both the robot and its load. Therefore, compliance plays a decisive role for a successful interaction between a robot and its environment.  
We choose to realize this compliance using Cartesian impedance control. 
Governed by the dynamics of a spring damper system, Cartesian impedance control exerts a wrench 

\begin{equation}  \label{eq:spring_damper_model}
	h = K_\mathrm{P}\left(\myx_\mathrm{o} - \myx_\mathrm{d}\right) + K_\mathrm{D} \left(\dot{\myx}_\mathrm{o}-\dot{\myx}_\mathrm{d}\right),
\end{equation}
according to the deviation of the object's actual pose $\myx_\mathrm{o}$ and twist $\dot{\myx}_\mathrm{o}$ from a desired, virtual set point  $\myx_\mathrm{d}$ and its twist $\dot{\myx}_\mathrm{d}$.
For the general case of six degree-of-freedom (DoF), all poses and twists are elements of the special Euclidean group \mbox{$\myx_\mathrm{o},\, \myx_\mathrm{d} \in SE\left(3\right)$} and \mbox{$\dot{\myx}_\mathrm{o}, \, \dot{\myx}_\mathrm{d} \in se\left(3\right)$}, respectively. The wrench \mbox{$h \in \mathbb{R}^6$} is a composite force/torque vector.
$K_\mathrm{P}$ and $K_\mathrm{D}$ are positive definite, diagonal gain matrices. We consider, without loss of generality that all of the above quantities are expressed w.r.t. a environment-fixed global frame of reference. 

Equation \eqref{eq:spring_damper_model} serves to illustrate the dynamical relation to a spring damper system, yet comes short of a thorough definition of stiffness in $SE\left(3\right)$. 
For a \textit{geometrically consistent} formulation of spring stiffness, we refer the reader to \cite{caccavale2008cartesianimpedance}. 

We assume a velocity-controlled set-point ~\mbox{$\dot{\myx}_\mathrm{d} = u \in \mathcal{U}$}.
Given a control trajectory~$u_{t_1 \rightarrow t_2} : [t_1, t_2] \rightarrow \mathcal{U}$, the evolution of the object's pose and twist can be computed using a standard physics simulator for rigid bodies with a six DoF spring-damper between set point and object as simulated impedance controller.

The pose and twist of the object are not exactly known and thus modeled as a joint probability distribution over poses and twists, denoted as the belief~$\mathrm{bel}\left(\myx_\mathrm{o},\, \dot \myx_\mathrm{o}\right)$.
As the dynamics of the object are deterministic, the evolution of the belief over time given a control trajectory is deterministic as well, assuming no measurements are used to update the belief.
Fig.~\ref{fig:figure_trajectory} depicts the evolution of the belief over the object's pose given an initial belief, a control trajectory and the dynamics of impedance-controlled rigid body motions.

The goal for our planner operating in this belief space is now to compute a control trajectory that brings a high fraction of the object's pose-belief close enough to the target pose.
In the next section a generalized formulation of a kinodynamic belief space planning problem is introduced.

\subsection{Kinodynamic Planning in Belief Space}\label{subsec:belief}

Let $x \in \mathcal{X}$ denote the state of our system. 
When dealing with  higher order systems, states may be composed of configurations as well as their respective derivatives, e.g. $x = (q,\, \dot{q})^\mathrm{T}$ for a second order process. 
At each time step $t$, state $x$ evolves according to the transition probability

\begin{equation}\label{eq:system_dynamocs}
	p \left(x' \mid x,\, u_{t \rightarrow t'}\right),
\end{equation}
 with \mbox{$x' \in \mathcal{X}$} being the system state at time \mbox{$t' = t + \Delta T$}, resulting from applying a control input \mbox{$u_{t\rightarrow t'}:\left[t, t+\Delta T\right] \rightarrow \mathcal{U}$}. 
Both the state space $\mathcal{X}$ and the control space $\mathcal{U}$ are assumed bounded.
 
A belief $s = \left(\mathrm{bel}\left(x\right),\, t\right)  \in \mathcal{BT} = \mathcal{B}\times  \left(0, \,T_\mathrm{max}\right]$ reflects  the knowledge about the system state at time $t$, where $\mathcal{B} = \left\lbrace \mathrm{bel}\left(x\right) |\, \mathrm{bel}\left(x\right) : \mathcal{X} \rightarrow \left[0, \, \infty\right) \right\rbrace$ is the set of all belief distributions and $T_\mathrm{max}$ is an upper bound on the time. 
Accordingly the belief state at time $t' = t+\Delta T$ is denoted $s' = \left(\mathrm{bel}'\left(x\right),\, t'\right)$. 
Note that the unusual notation of a belief state as a tuple in $belief\times time$ space aims to facilitate the later application of kinodynamic planning algorithms, wherein time is usually treated as a separate dimension. 
A belief state $s$  evolves according to the Bayesian laws of probability as

\begin{equation}
	s' = \left( \gamma\, p\left(z'\mid x'\right) \, \int_{\mathcal{X}} \mathrm{bel}\left(\hat{x}\right)\,p \left(x' \mid \hat{x},\, u_{t \rightarrow t'}\right)\, \mathrm{d}\hat{x},\, t'\right),
\end{equation}
where $\gamma$ is a normalization factor and $z'$ is the measurement at time $t'$. 
We omit the measurement update with regard to the remainder of this work for the following reasons: 
the scope of our applications are target environments with highly non-linear, constrained and hybrid system dynamics for which the formulation of an adequate  measurement update constitutes a major challenge. 
Furthermore, as a prerequisite of the planning algorithm presented in this paper, we require the evolution of the belief to be deterministic, a property that is lost when incorporating stochastic measurements. Note that in order to maintain determinism, one could employ approximative techniques such as maximum likelihood measurements \cite{platt2010belief}. 
The resulting belief space dynamics are written as
\begin{equation}
	s' = \left( \int_{\mathcal{X}} \mathrm{bel}\left(\hat{x}\right)\,p \left(x' \mid \hat{x},\, u_{t \rightarrow t'}\right)\, \mathrm{d}\hat{x},\, t'\right).
\end{equation}

What follows is the formal definition of our target, the generation of optimal belief space trajectories.

Let $s_\mathrm{start} = \left( \mathrm{bel}\left(x_\mathrm{start}\right),\,  t_\mathrm{start}\right)$ be the initial configuration in $ belief\times time$ space and let furthermore $\mathcal{F} \subset \mathcal{BT}$ be the set of all valid belief states.

\begin{defn} \label{dfn:valid}
  	 A trajectory $\tau : t \in \left[t_\mathrm{start},\, t_\mathrm{stop}\right] \mapsto \tau\left(t\right) = \left(\mathrm{bel}\left(x\right),\, t\right)$ is said to be \textit{valid} iff it lies entirely within $\mathcal{F}$ and is generated in accordance to the system dynamics \eqref{eq:system_dynamocs}, by a control sequence $u:\left[t_\mathrm{start},\, t_\mathrm{stop}\right] \rightarrow \mathcal{U}$. 

\end{defn}

\begin{rema}
	A region in $ belief\times time$ space $\mathcal{R} \subset \mathcal{BT}$  is a measure of probability mass and not to be confused with a region in state space $\mathcal{X}$. Therefore it is possible to employ parametric probability density functions with infinitely long tails. For instance, one could declare the \textit{valid} region \mbox{$\mathcal{F} \subset \mathcal{BT}$} to be the region where more than \unit[95]{\%} of the probability mass is formed by states, in which the interaction forces are below a distinct upper bound.
\end{rema}

Given an end-game region $\mathcal{G} \subset \mathcal{BT}$  allows us to define a \textit{feasible} trajectory.

\begin{defn}\label{dfn:feasible}
	A trajectory $\tau\left(t\right)$ with $t \in \left[t_\mathrm{start},\, t_\mathrm{stop}\right]$ is considered \textit{feasible} iff it is valid, $\tau\left(t_\mathrm{start}\right) = s_\mathrm{start}$ and   $\tau\left(t_\mathrm{stop}\right) \in \mathcal{G}$.
\end{defn}

The quality of a trajectory $\tau : t \in \left[t_\mathrm{start},\, t_\mathrm{stop}\right]$ is measured with the cost function
\begin{equation} \label{eq:cost}
	C(\tau) = \int_{t_\mathrm{start}}^{t_\mathrm{stop}} \ell \left(\tau\left(\nu\right)\right) \mathrm{d}\nu,
\end{equation}
where $\ell\left(s\right) : s \in \mathcal{BT} \rightarrow [l_\text{min}, l_\text{max}]$ and~$l_\text{min} > 0$, is the immediate cost experienced in state $s$. 
Optimality with respect to this cost function is what defines an \textit{optimal} trajectory.

\begin{defn}\label{dfn:optimal}
A trajectory $\tau^*\left(t\right)$ is \textit{optimal} iff it is \textit{feasible} and minimizes
\[ C\left(\tau^*\right) = \underset{\tau}{\min}\ C\left(\tau\right).\]
\end{defn}

\vspace{0.1cm}
\section{Optimal Belief Space Planning} \label{sec:planning}

Our approach to optimal belief space planning makes use of the work of Hauser and Zhou~\cite{hauser_aox} wherein the AO-$x$ meta-algorithm is presented.
This algorithm uses a probabilistically complete, kinodynamic planner repeatedly to plan optimal in the limit.
To achieve optimal belief space trajectories we introduce a probabilistically complete, kinodynamic belief space planner in Section~\ref{subsec:planner_1} and then use it in conjunction with the AO-$x$ meta-algorithm for optimal planning in Section~\ref{subsec:planner_2}.

\subsection{Kinodynamic Belief Space Planning}\label{subsec:planner_1}

To allow kinodynamic planning in a belief space, we adapt the EST planner~\cite{est_moving_obstacles_icra}, a kinodynamic motion planner, to probabilistic domains.
We therefore call our planner Belief-EST (\textbf{B-EST}).
\begin{algorithm}[h]
\begin{algorithmic}\label{alg:algorithm_est}
	\Procedure{\textbf{algorithm B-EST}$\left(s_\mathrm{start},\, \mathcal{G}, \, c_\mathrm{best}\right)$ }
	\vspace{0.1cm} 
	\State $V$ = $\left\lbrace s_\mathrm{start} \right\rbrace,\ E$ = $\left\lbrace \right\rbrace$ 
	
	\While {withinTimeBudget$\left(\right)$}
	
	\State $s$ = sampleWeighted$\left(V\right)$
	\State $u_{t\rightarrow t'}$ = sampleControl$\left(\right)$  
	\State $s'$ = integrate$\left(s,\,u_{t\rightarrow t'}\right)$

	\If {isValid$\left(s,\, u_{t\rightarrow t'} ,\, c_\mathrm{best}\right)$}
		\State $V$.append$\left(s'\right)$
		\State $E$.append$\left(s,\, u_{t\rightarrow t'} ,\, s'\right)$
		
		\If {$s'\in \mathcal{G}$}
		\State \Return trajectory from $s_\mathrm{start}$ to $s'$ 
		\EndIf
	\EndIf
	\EndWhile
	\State \Return failure
	\EndProcedure
\end{algorithmic}
\end{algorithm}
Our algorithm constructs a tree rooted at the initial state $s_\mathrm{start}$. 
Each iteration starts off with sampling a node from the already constructed tree via the procedure \textbf{sampleWeighted}. 
This procedure aims to select nodes from the tree in a way that prefers sparsely covered areas of the belief space. 
Next, the algorithm samples a piecewise linear control input $u_{t\rightarrow t'}$ and according to \eqref{eq:system_dynamocs}, integrates the system dynamics. Sampling control inputs at random allows to directly enforce actuation constraints and spares the intricate design of a steering function. 
The \textbf{isValid} function rejects an expansion towards $s'$, if the segment from $s$ to $s'$ is \textit{invalid} according to Definition \ref{dfn:valid}. Furthermore, it discards any state $s'$ that  exceeds an upper cost-bound $c_\mathrm{best}$. 
If $s'$ is contained in the end-game region $\mathcal{G}$, the algorithm terminates, returning the feasible trajectory.

\subsection{Optimal Belief Space Planning}\label{subsec:planner_2}
The $\text{AO-}x$ meta algorithm~\cite{hauser_aox} is a generic optimality wrapper for kinodynamic planners such as the above introduced B-EST algorithm. First, B-EST is augmented by a separate cost dimension. Thereafter,  $\text{AO-}x$ achieves optimality by repeatably calling the \textit{feasible} B-EST planner while lowering the upper cost dimension bound.
\begin{algorithm}[h]
\begin{algorithmic}
	\Procedure{\textbf{algorithm AO-B-EST}$\left(s_\mathrm{start},\, \mathcal{G}\right)$} 
		\State $\tau$ = B-EST$\left(s_\mathrm{start},\, \mathcal{G},\, \infty\right)$
		\State $c_\mathrm{best}$ = cost$\left(\tau\right)$
	\While{withinTimeBudget$\left(\right)$}
		\State $\tau$ = $\text{B-EST}\left(s_\mathrm{start},\, \mathcal{G},\, c_\mathrm{best}\right)$
		\State $c_\mathrm{best}$ = cost$\left(\tau\right)$
	\EndWhile
	\State \Return $\tau$
	\EndProcedure
\end{algorithmic}
\end{algorithm}

Let us briefly outline the concrete implementation AO-B-EST. 
In a first step B-EST is queried for an initial trajectory.
Next, B-EST is repeatedly queried, with its expansion being constraint to produce results with solution costs, less than the current cost bound $c_\mathrm{best}$. 
Once exceeding a given time budget, the algorithm terminates returning the best solution found.

\subsection{Completeness and Optimality}

In order to implement a belief space planner on a computer, the representation of probability distributions must be achieved with a finite set of parameters.
Let~$\mathrm{bel}\left(x;\,  \theta\right)$ be the belief over a state~$x$ represented by a vector of parameters~$\theta \in \Theta$.
An example for a parameter vector~$\theta$ would be the data, stored in a particle cloud. Given a state represented by~$n$ values and a particle cloud consisting of~$N$ particles, the parameter vector~$\theta$ would be of  dimension~$\mathrm{dim}\left(\theta\right)=n\,N$.

As we assume a deterministic evolution of the belief, the planning problem in belief space can be transformed to a kinodynamic planning problem in belief-parameter space.
The initial state~$\theta_\mathrm{start}$ of this planning problem represents the parametrization of the initial belief~\mbox{$s_\mathrm{start} = (\mathrm{bel}\left(x;\,  \theta_\mathrm{start}\right),\, t_\mathrm{start})$}.
This initial parameter state must be transformed into a final state within~\mbox{$\Theta_\text{goal} = \{\theta \in \Theta | \mathrm{bel}\left(x;\, \theta\right) \in \mathcal{G}\}$} 
via the parameter dynamics~$\dot\theta = f(\theta,\, u)$.
Costs and valid parameter spaces can be defined analogously.
For a parametrized belief space, our planner B-EST is therefore equivalent to the EST~\cite{est_moving_obstacles_icra} planner in belief parameter space.

Under the assumption of $\alpha$-$\beta$-Expansiveness~\cite{est_moving_obstacles_journal} of the belief parameter space and a non-zero relative volume of the reachable set of goal parameters, B-EST is probabilistically complete.
This follows directly from the equivalence of \mbox{B-EST} in belief space to EST in belief-parameter space and the proof of~\cite{est_moving_obstacles_journal}.
The question remains whether the belief space of our assembly domain fulfills this expansiveness property.
Our experiments and their results in Section~\ref{results} strongly indicate that this is the case.

In~\cite{hauser_aox} the asymptotic optimality of the AO-$x$ algorithm is proven under the assumption of a probabilistically complete kinodynamic planner and an additional well-behavedness condition.
From this follows that under the assumption of an expansive belief-parameter space and the well-behavedness condition of~\cite{hauser_aox}, AO-B-EST is asymptotically optimal.

\section{Planning Compliant Motions for Assembly} \label{sec:planning_for_assembly}
This section describes the steps to be taken in order to make the previously introduced planning methods accessible to assembly tasks.

 Having outlined our realization of compliant motions in Section~\ref{subsec:compliance}, we can now introduce the parameters $\theta$ of an approximate belief state representation $\mathrm{bel}\left( x; \, \theta \right)$ suitable for assembly planning. We choose to approximate the belief state by a set of $N$ particles, where each particle $p_i = \left(\myx_\mathrm{o}^i,\, \dot{\myx}_\mathrm{o}^i\right)$ represents the pose and twist of the object.
 
 The particles' initial distribution is generated by adding Gaussian noise to the grasp transform that is, the transform from the center of the object to the robot end-effector. This aims to represent prominent sources of uncertainty such as localization errors whilst grasping the object. Further sources of uncertainty could include motion noise or a noise prone manipulator pose e.g. when considering mobile manipulators. 
 
Each of the particles is controlled via an impedance 
\begin{equation}\label{eq:springdamper}
	h_i =  K_\mathrm{p}\,\left(\myx^i_\mathrm{o}-\myx_\mathrm{d}\right) + K_\mathrm{d}\,\left(\dot{\myx}^i_\mathrm{o} - \dot{\myx}_\mathrm{d}\right).
\end{equation}
However, all particles share a \textit{common} reference $ \myx_\mathrm{d}, \dot{\myx}_\mathrm{d}$. Consequently, the belief parameter vector $\theta$ is chosen as
\[\theta = \left(p_1,\, p_2,\, \dots,\, p_N,\, \myx_\mathrm{d},\, \dot{\myx}_\mathrm{d}\right)^\mathrm{T}.\]
	
Depending on the problem setting it can be advantageous to project the particles and their reference trajectory into subspaces of $SE\left(3\right)$, resulting in a planning space of $N\dim\left(p\right)+ \dim\left( \myx_\mathrm{d}\right)+ \dim\left( \dot{\myx}_\mathrm{d}\right)$ dimensions. Fig.~\ref{fig:figure_trajectory} depicts our approach of planning a set-point trajectory for a belief state represented as a particle cloud.

The immediate cost $\ell\left(\theta\right)$ is defined as the average over the immediate cost of each individual particle
\[
\ell\left(\theta\right)= \frac{1}{N}\sum_{i=1}^{N} \ell\left(p_i\right),
\] 
where $\ell\left(p_i\right)$ denotes the contribution in cost of each particle. Since we seek to 
minimize the use of excessive forces and torques during the assembly process, we penalize any motion that further stretches the virtual spring damper system. The immediate cost is defined as the work that is attributed towards stretching the spring and is calculated as $\ell\left(p_i\right) = -\frac{1}{2}\left(1-\mathrm{sgn}\left(h^{i,\mathrm{T}}\,\dot{\myx}^i_\mathrm{o}\right)\right)h^{i,\mathrm{T}}\,\dot{\myx}^i_\mathrm{o}$. With this choice of an immediate cost, the trajectory cost $C\left(\tau\right)$ defined in~\eqref{eq:cost} on average reflects the energy spent towards undesired, forceful assembly motions. Note that this choice of energy cost fortunately  bypasses the need to weigh off rotational and translational quantities. 

Let us briefly describe the remaining components of the above covered algorithms. 
The $\textbf{sampleWeighted}$ method aims to select states that quickly expand the tree towards sparsely covered areas, avoiding culminations that stem from repeatedly expanding from nodes in the vicinity of the tree's root. Literature provides several methods to facilitate a quick expansion towards a target node (e.g. \cite{phillips2004guided}).
 We choose to group all tree nodes into cells of a coarse discretization over a sub-manifold of the planning space, namely, the mean posture of the particle cloud.

 The method then returns a random node from a randomly selected, occupied cell. The B-EST algorithm's $\textbf{integrate}$ method propagates each particle given a control command. We choose the input as the reference velocity of the manipulator $u = \dot{\myx}_\mathrm{d}$ and furthermore constrain it to be piecewise linear. During the integration step, a node's particle is removed 
whenever the $L_2$ norm of its end-effector wrench $h_i$ exceeds a hardware dependent threshold. The $\textbf{isValid}$ function prohibits the tree's expansion towards a node that either exceeds the current cost barrier, or in the case that too many of the particles have been removed during the integration step. The particle approximation of the belief state is considered to be contained within the end-game region $\mathcal{G}$ if more than a $\gamma$-fraction of the particles is contained within a ball $\mathcal{B}_r  = \left\lbrace \myx_\mathrm{o}^i\, |\, d\left(\myx_\mathrm{o}^i - \myx_\mathrm{goal}\right) < r  \right\rbrace$, where $d \left(\cdot\right) \rightarrow \mathbb{R}^+$ denotes a suitable distance function to a goal pose $\myx_\mathrm{goal}$.

\section{Results}
\label{results}
The following section introduces a set of benchmark problems with regard to which we evaluated our planner (Section~\ref{subsec:benchmarks}) and gives details on our implementation (Section~\ref{subsec:impl}). We analyze key properties such as the cost-convergence or the planner's ability to find solutions in appropriate time in Section~\ref{subsec:resultsanalysis}. Section~\ref{subsec:resultsreal} interprets the results gathered from thorough real-world experiments. 

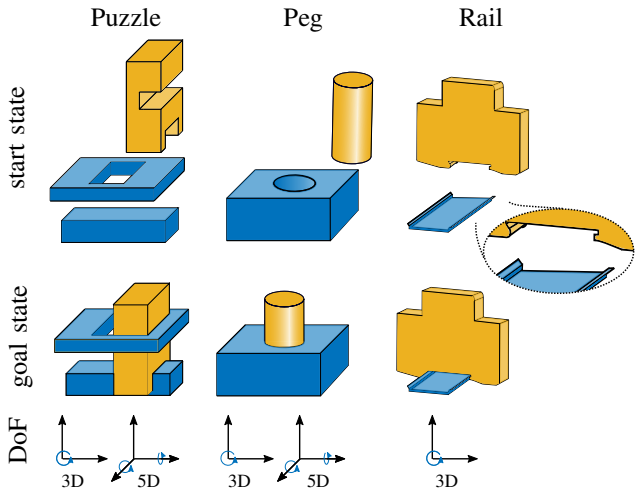
\begin{figure}
	\vspace{0.7cm}
	\centering
	\begin{overpic}[width=0.9\linewidth]{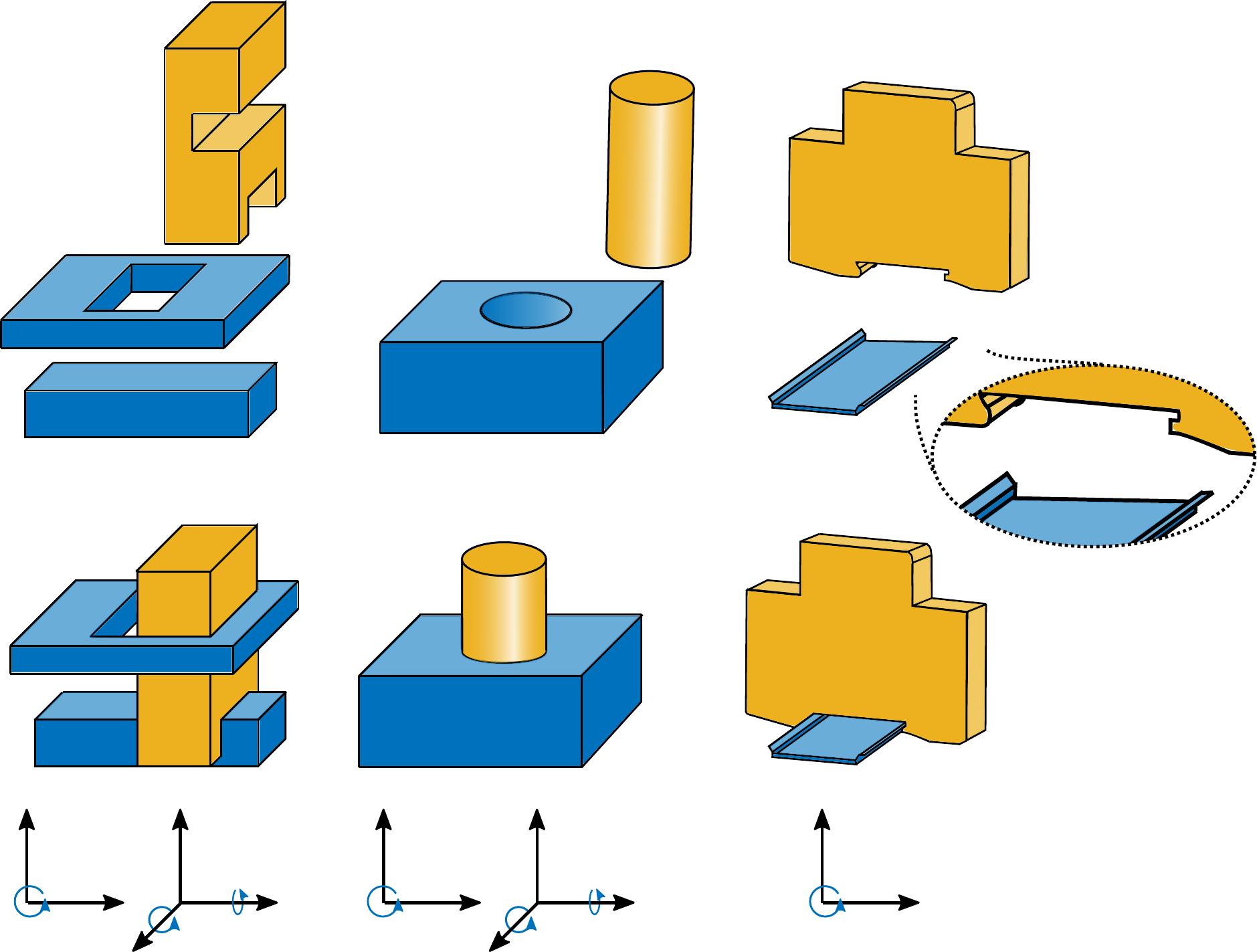}
		\put(70,78.00){{Rail}}	
		\put(40,78.00){{Peg}}	
		\put(7,78.00){{Puzzle}}	
		
		\put(-7,16.00){\rotatebox{90}{{{goal state}}}}
		\put(-7,3.00){\rotatebox{90}{{{DoF}}}}
		\put(-7,50.00){\rotatebox{90}{{{start state}}}}
		\put(2,-1.00){\scriptsize{3D}}	
		\put(15,-1.00){\scriptsize{5D}}
		\put(30.5,-1.00){\scriptsize{3D}}	
		\put(44,-1.00){\scriptsize{5D}}	
		\put(66,-1.00){\scriptsize{3D}}

	\end{overpic}
	\caption{Three benchmark problems: the puzzle problem requiring three consecutive joining motions, the well studied peg-in-hole problem and an assembly task that requires mounting an electrical fuse onto a rail.}
	\label{fig:benchmarks}
	\vspace{-0.3cm}
\end{figure}
\subsection{Benchmark Problems}
\label{subsec:benchmarks}

We evaluated our approach on three notably distinct benchmark problems, depicted in Fig.~\ref{fig:benchmarks}. 
Before moving on to a detailed description of the individual benchmarks, it is important to stress the following: each of the problems was solved using a common version of the algorithm.
Neither did we require problem specific modifications of the cost function nor was it necessary to specify directions or surfaces of constraint or free motion. 
Input was given in the form of a description of the geometry (CAD files), the initial end-effector pose and a goal condition. 
We assumed the grasp transform to be noise prone. We chose a translational standard deviation of \unit[2.5]{mm} and a rotational standard deviation of \unit[0.015]{rad}. 
The physics simulation including collision constraint resolution and the active compliant end-effector, was carried out in $SE\left(3\right)$. 
The rotational and translational spring stiffness were chosen as $k_\mathrm{P,trans} = $ \unitfrac[1000]{N}{m} and \mbox{$k_\mathrm{P,rot} = $ \unitfrac[60]{Nm}{rad}}. 
The damping coefficients were chosen such that the Cartesian impedance is aperiodically damped. 
We restricted the planning domain to down-projections of three and five DoF respectively. 

\begin{itemize}
	\item Puzzle: The puzzle problem (taken form \cite{dakin1992simplified}) requires three consecutive mating motions. The tolerances lie within \unit[1]{mm} - \unit[2]{mm}.
	\item Peg: The peg-in-hole is a well studied problem since it covers a wide spectrum of assembly tasks found in industrial production. The tolerance is chosen as~\unit[0.5]{mm}. The hole is only \unit[5]{\%} wider that the peg.
	\item Rail: The rail problem requires mounting a fuse onto a top hat rail as typically found in industrial switch cabinets. The fit exhibits zero clearance, which in face of the above chosen uncertainties, constitutes a major challenge. This task aims to show our method's relevance to industrial applications.
\end{itemize}

\subsection{Implementation} \label{subsec:impl}
We simulated the compliant end-effector and its environment using the Bullet  physics engine \cite{coumans2013bullet}. The particle dynamics were propagated in parallel using separate physics engine instances. We ran our experiments on a ten-core Intel Xeon E5-2650v3.
 
For the real-world experiments we used a KUKA iiwa 7 R800 redundant 7-axis manipulator with joint-torque feedback, interfaced at a high-level command rate of \unit[200]{Hz}. 
The stiffness and damping values of the robot's Cartesian impedance mode were chosen to match the ones used during the planning phase. 
The reference frame for the Cartesian impedance was set to match the object frame $\myx_\mathrm{o}$.

\subsection{Results - Convergence}
\label{subsec:resultsanalysis}

\begin{figure}
	\vspace{0.2cm}
	\hspace{0.7cm}
	\centering
	\begin{overpic}[width=0.9\linewidth]{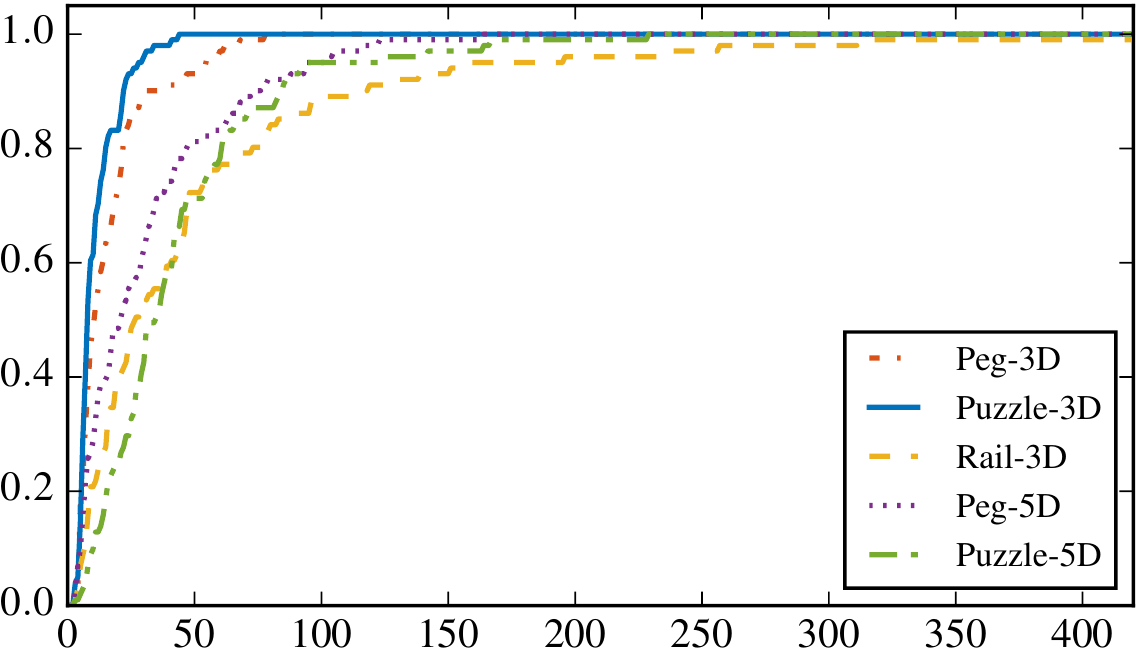}
		
		\put(40,-5.00){\footnotesize{{ planning time $\left[\mathrm{s}\right]$}}}
		\put(-6,20.00){\rotatebox{90}{\footnotesize{{success rate}}}}
		
	\end{overpic}
	\vspace{0.1cm}
	\caption{Success rates of B-EST for all five benchmarks. Each line visualizes the success rate of 100 planner calls for each of the benchmark tasks.}
	\label{fig:figure_solution_found}	
\end{figure}

An important property of a planner is its ability to find a solution in finite time. 
Fig.\ref{fig:figure_solution_found} depicts the planner's success rate with regard to planning time. 
 For each of the benchmarks, we conducted 100 queries with a maximum planning time of \unit[420]{s} and $N=10$ particles. 
The results clearly demonstrate the planner's ability to quickly come up with solutions.

\begin{figure}
	\hspace{0.7cm}
	\centering
	\begin{overpic}[width=0.9\linewidth]{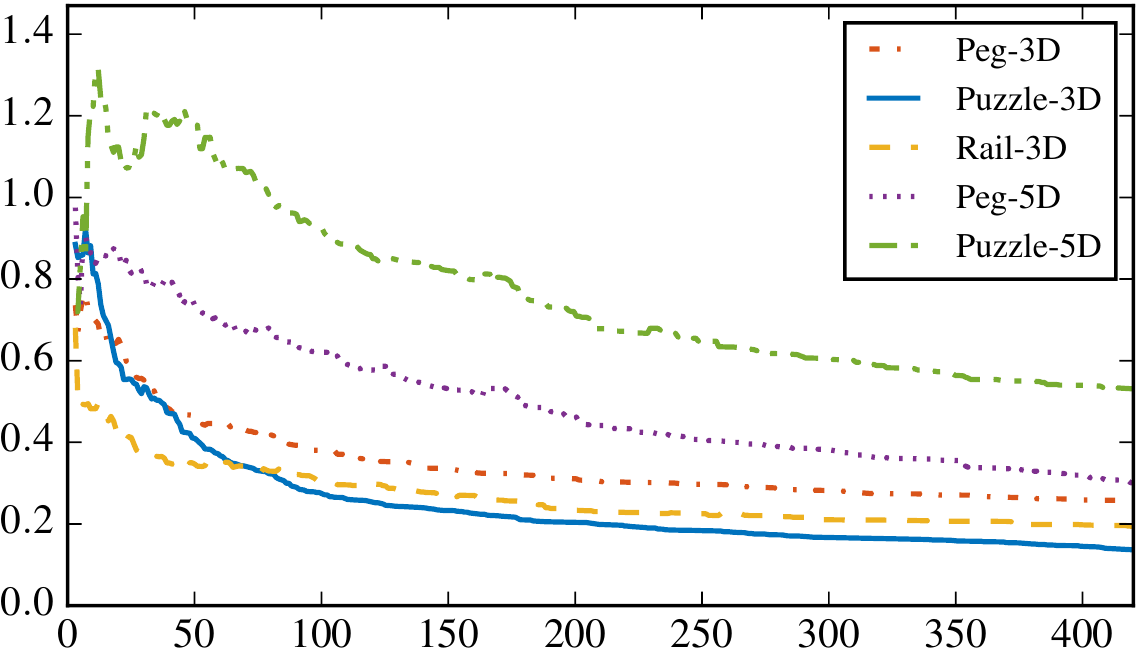}
		
		\put(40,-5.00){\footnotesize{{ planning time $\left[\mathrm{s}\right]$}}}
		\put(-6,25.00){\rotatebox{90}{\footnotesize{{cost}}}}
		
	\end{overpic}
	\vspace{0.1cm}
	\caption{Mean costs of the five benchmark tasks. Each line visualizes the average cost of successful runs for one benchmark of 100 planner calls.}
	\label{fig:figure_average_cost}	
\end{figure}

Fig.~\ref{fig:figure_average_cost} illustrates the average cost of successful assembly trajectories over time. The cost function employed is given in Section~\ref{sec:planning_for_assembly}. Again, we carried out 100 consecutive experiments with a maximum planning time of \unit[420]{s}. Our results show the asymptotic decay of cost over time, which is in strong accordance to the experiments and results of \cite{hauser_aox}. It can be seen that the problems of higher dimensionality produced higher costs. This behavior is related to the fact that problems in higher dimensional spaces require more planning time and hence undergo a smaller amount of improvement iterations.

\begin{figure}[h]
	\vspace{0.2cm}
		\hspace{0.7cm}
		\centering
		\begin{overpic}[width=0.9\linewidth]{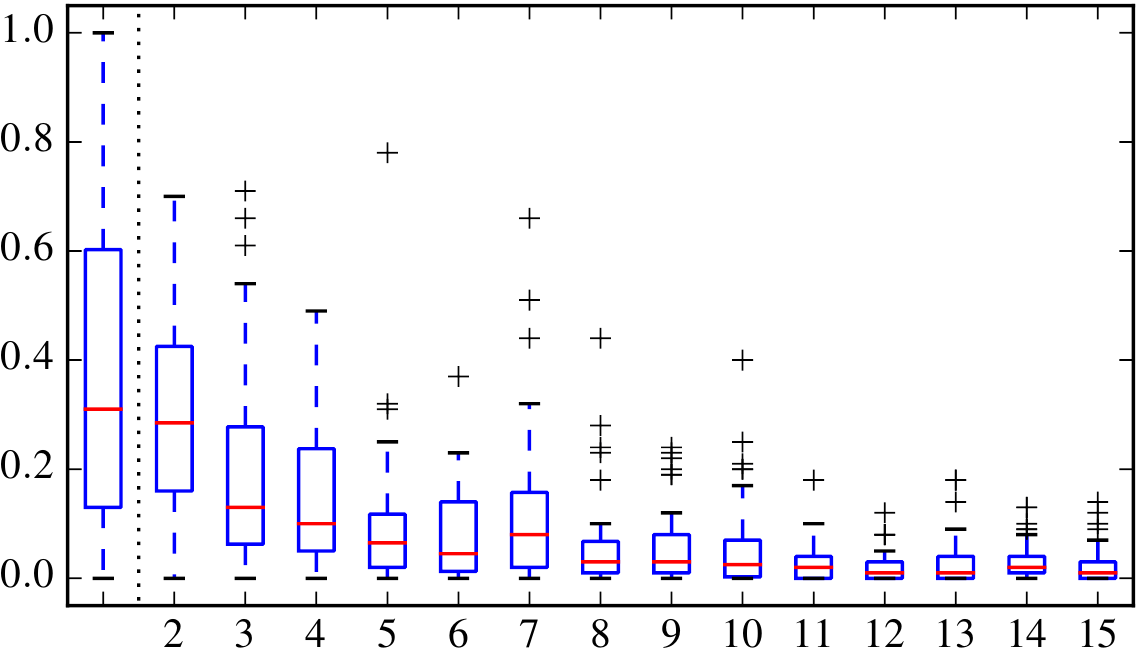}

			\put(3,1.40){\tiny{\textsf{ \color{red}{Standard}}}}
			\put(6.1,-1.7){\tiny{\textsf{ \color{red}{EST}}}}

			\put(40,-5.00){\footnotesize{{number of particles $N$}}}
			\put(-6,22.00){\rotatebox{90}{\footnotesize{{failure rate}}}}
		\end{overpic}
		\vspace{0.1cm}
		\caption{A motion plan (Peg-3D) is computed for a sample set containing $N$  particles. Each of the plans is evaluated in 100 consecutive experiments, with initial states drawn from the same distribution as used during the planning phase. The averaged results (50 repetitions) emphasize the effectiveness of our approach.}
		\label{fig:figure_particles}	
		\vspace{-0.2cm}
\end{figure}
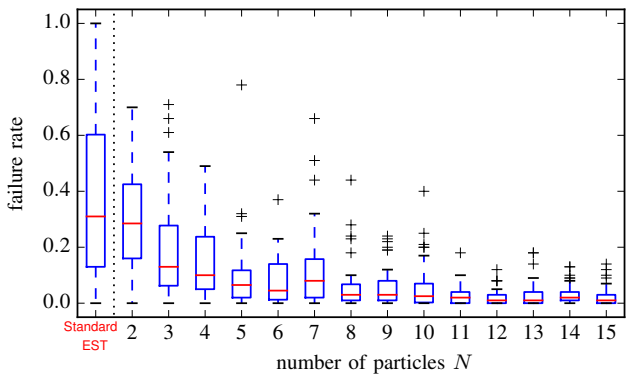

Finally, we evaluated the number of particles required in order to produce robust and successful assembly trajectories. The experiment was conducted as follows. We computed assembly trajectories for the peg-in-hole benchmark problem, increasing the amount of  particles $N$. The resulting trajectories were evaluated in 100 consecutive experiments with initial states drawn from the same distribution as used during the planning phase. We repeated this plan-evaluate procedure 50 times. The averaged failure rates are depicted in Fig.~\ref{fig:figure_particles}. The mean failure rate of the standard EST approach is as high as \unit[31]{\%}. By contrast, we were able to achieve an average failure rate of \unit[1]{\%} with twelve particles only. This result shows the gain in robustness by our particle-based approach to parts assembly. 

\subsection{Results - Real World Experiments} \label{subsec:resultsreal}

We evaluated both, our AO-B-EST with $N=10$ particles and an EST planner based on the Peg-5D, the Puzzle-5D and the Rail-3D benchmark problems. We computed 14 individual trajectories for each of the benchmarks. Each of the trajectories was then evaluated in 5 consecutive experiments. In total, we conducted 420 real-world assembly experiments.

The experiments proceeded as follows: we detect the object using a wrist-mounted camera in combination with industrial grade software for object pose measurement. The manipulator then grasps the object using a two-jaw gripper, transfers it to the mating object and finally executes the assembly plan. The employed setup is well calibrated and thus far more precise than assumed during the planning phase. In order to recreate the grasp uncertainty we distorted the system's knowledge about the grasp pose using the same Gaussian noise as assumed during planning.

\begin{table}[h]
	\caption{Success rates in real world experiments}
	\label{table_example}
		\begin{tabular}{ l|cc|cc|cc }   
		    \toprule
			benchmark & \multicolumn{2}{c|}{Peg-5D} & \multicolumn{2}{c|}{Rail-3D} &  \multicolumn{2}{c}{Puzzle-5D} \\
			\midrule
			planner & \rotatebox[origin=c]{90}{{EST}} & \rotatebox[origin=c]{90}{{\hspace{5pt}AO-B-EST}} & \rotatebox[origin=c]{90}{{EST}} & \rotatebox[origin=c]{90}{{\hspace{5pt}AO-B-EST}} & \rotatebox[origin=c]{90}{{EST}} & \rotatebox[origin=c]{90}{{\hspace{5pt}AO-B-EST}} \\ 
			\midrule
			success rate    & \textbf{67\%} & \textbf{96\%} & \textbf{51\%} & \textbf{90\%} & \textbf{54\%} & \textbf{90\%}   \\ 
			\midrule
			p-value (70 samples) & \multicolumn{2}{c|}{$1.63\times10^{-5}$} & \multicolumn{2}{c|}{$6.81\times10^{-7}$} & \multicolumn{2}{c}{$3.40\times10^{-6}$}\\ 
			Fisher's exact test& \multicolumn{2}{c|}{} & \multicolumn{2}{c|}{} & \multicolumn{2}{c}{}\\ 
			\midrule
			p-value (14 means)& \multicolumn{2}{c|}{$1.90\times10^{-2}$} & \multicolumn{2}{c|}{$9.06\times10^{-5}$} & \multicolumn{2}{c}{$3.99\times10^{-4}$}\\ 
			Welch's t-test& \multicolumn{2}{c|}{} & \multicolumn{2}{c|}{} & \multicolumn{2}{c}{}\\ 

			\bottomrule
		\end{tabular}
\end{table}




\textsc{table} \ref{table_example} shows the success rates for both AO-B-EST and an EST planner. 
Despite the artificially added measurement uncertainty and the gap between simulation and reality, AO-B-EST yielded robust assembly trajectories throughout our experiments. 
During the planning phase, we allowed for a maximum end-effector force of \unit[30]{N} and a maximum torque of \unitfrac[3]{Nm}{rad}, respectively. 
With the purpose of reducing uncertainty about the object's state, \mbox{AO-B-EST} tends to fully exploit this wrench threshold, which occasionally led to a slippage of the grasped object.
This violates the assumption of rigidly grasped objects.
We assume that modeling this slippage will bring the success rates closer to 100\%. 
To evaluate robustness and compare the two planners two different scenarios must be addressed separately: on-line planning and off-line planning.

In the on-line planning scenario the trajectories for assembly are computed immediately before assembly and are used only once.
To compare the two planners we use Fisher's exact test on the full data set of 70 trials for each benchmark task and test against the null-hypothesis of equal success-rates.
This analysis shows that our planner significantly (at 1\% p-value) outperforms the baseline.
However, the 70 samples for each pair of planner and benchmark are not independent as groups of five samples are produced by the same trajectory sample, which may bias the result.

In the off-line planning scenario an assembly trajectory is computed off-line and then used for multiple products during execution.
For this case the average success-rate is a random variable dependent on the sampled trajectory.
To compare both planners in the off-line case we compute the average success-rate of each trajectory which yields 14 data-points per pair of planner and benchmark, each computed with five real world experiments.
We analyze the resulting data-set under the assumption of independent Gaussian distributions using Welch's unequal variance t-test and test against the null-hypothesis of equal average success-rates.
Again, this analysis shows that our planner significantly~(at most 2\% p-value) outperforms the baseline.

\section{CONCLUSION AND FUTURE WORK} \label{sec:conclusion}

In this paper we introduce a novel approach to model robotic assembly tasks under uncertainty and under the full complexity of non-linear contact-dynamics.
This is achieved by formulating a belief space planning problem over an impedance-controlled system.
To solve this problem class we propose a randomized belief space planner that is asymptotically optimal.
The planner operates over a particle-based representation of uncertainty and utilizes physics simulation to model the spring-damper dynamics of the underlying impedance-controlled robot.

As our algorithm uses a control-based system model, computed trajectories can be directly executed on the real robot without post-processing.
Thorough real-world experiments demonstrate that our planner enables reliable assembly solely based on a CAD (geometry, mass, inertia) description of the respective parts.
An important qualitative result is that the solutions produced by our planner include motions that actively reduce uncertainty in the assembly process.

Our approach is currently limited to a model of the assembly process that does not include sensor-feedback besides the active impedance control. Promising avenues for future research include the extension of our modeling-approach to a full POMDP, the inclusion of increasingly complex dynamics such as deformable objects and a thorough comparison to results from related state-of-the-art planners.

\addtolength{\textheight}{-7.0cm}

\bibliographystyle{./IEEEtran} 
\bibliography{./IEEEabrv,./IEEEexample}

\end{document}